\documentclass[10pt,twocolumn,letterpaper]{article}

\usepackage[pagenumbers]{cvpr} % To force page numbers, e.g. for an arXiv version

\usepackage{blindtext}
\usepackage{graphics}
\usepackage{caption}
\usepackage{colortbl}
\definecolor{mygray}{RGB}{234,234,234}
\usepackage{color}
\usepackage{xcolor}
\usepackage{graphicx}
\usepackage{booktabs}
\usepackage{amsmath, amsthm, amsfonts, amssymb}
\usepackage{mathrsfs}
\usepackage{textcomp}
\usepackage{epstopdf}
\usepackage{multirow}
\usepackage{wrapfig}
\usepackage{booktabs} % For formal tables
\usepackage[ruled]{algorithm2e} % For algorithms
\usepackage{algorithmicx}  
\usepackage{algpseudocode}
\usepackage{ragged2e}
\usepackage[normalem]{ulem}
\usepackage{bm}
\usepackage{ulem}
\usepackage{makecell}
\usepackage{diagbox}

\SetAlFnt{\small}
\SetAlCapFnt{\small}
\SetAlCapNameFnt{\small}
\SetAlCapHSkip{0pt}

\definecolor{green}{rgb}{0, 0.5, 0}
\definecolor{orange}{rgb}{0.6, 0.3, 0.1}
\definecolor{red}{rgb}{1.0, 0.0, 0.0}
\definecolor{teal}{rgb}{0.0, 0.4, 0.4}
\definecolor{purple}{rgb}{0.65,0,0.65}
\definecolor{saffron}{rgb}{0.95,0.75,0.2}
\definecolor{turquoise}{rgb}{0.0,0.5,0.5}
\definecolor{brown}{rgb}{0.5, 0.16, 0.16}
\definecolor{brickred}{rgb}{.6, .2 .1}
\definecolor{coral}{rgb}{1,0.45,0.33}
\definecolor{newcolor}{rgb}{.8,.349,.1}

\definecolor{cvprblue}{rgb}{0.21,0.49,0.74}
\usepackage[pagebackref,breaklinks,colorlinks,citecolor=cvprblue]{hyperref}

\def\confYear{2024}

\title{EmoGen: Emotional Image Content Generation with \\ Text-to-Image Diffusion Models }

\begin{document}

\author{Jingyuan~Yang,
	Jiawei~Feng,
	Hui~Huang\footnotemark[1]\\
	Shenzhen University \\	
	{\tt\small \{jingyuanyang.jyy, fengjiawei0909, hhzhiyan\}@gmail.com}
	\vspace{-25pt}
}

%\author{Jingyuan~Yang\\
%	Shenzhen University\\
%	{\tt\small jingyuanyang.jyy@gmail.com}
%	% For a paper whose authors are all at the same institution,
%	% omit the following lines up until the closing ``}''.
%% Additional authors and addresses can be added with ``\and'',
%% just like the second author.
%% To save space, use either the email address or home page, not both
%\and
%Jiawei~Feng\\
%Shenzhen University\\
%{\tt\small fengjiawei0909@gmail.com}
%\and
%Hui~Huang\thanks{Corresponding author}\\
%Shenzhen University\\
%{\tt\small hhzhiyan@gmail.com}
%\vspace{-25pt}
%}

\twocolumn[{
	\renewcommand\twocolumn[1][]{#1}
	\maketitle
	\begin{center}
		\centering
		\includegraphics[width=0.95\linewidth]{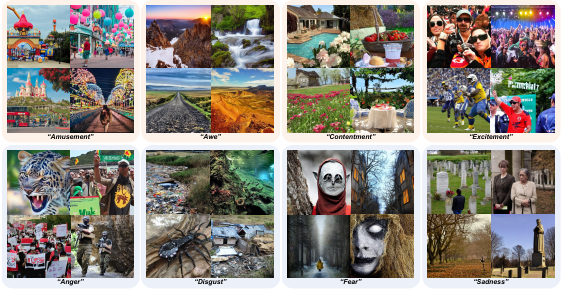}
		%		(a)\hspace{32mm}(b)\hspace{32mm}(c)\hspace{32mm}(d)\hspace{32mm}(e)
%			\vspace{-5pt}
		\captionof{figure}{Emotional Image Content Generation (EICG). Given an emotion category, our network produces images that exhibit unambiguous meanings (\textit{semantic-clear}), reflect the intended emotion (\textit{emotion-faithful}) and incorporate varied semantics (\textit{semantic-diverse}).}
%			Experimental results of EmoGen. (a) represents the diverse semantic contents under each emotion and (b) indicates emotional semantics can be fused to neutral objects/scenes, smoothly and emotionally.}
			\vspace{0pt}
		\label{fig:teaser}
	\end{center}
}]

\renewcommand{\thefootnote}{\fnsymbol{footnote}}
\footnotetext[1]{Corresponding author}

\begin{abstract}
	
\vspace{-10pt}
	Recent years have witnessed remarkable progress in image generation task, where users can create visually astonishing images with high-quality.
	However, existing text-to-image diffusion models are proficient in generating concrete concepts (dogs) but encounter challenges with more abstract ones (emotions).
	Several efforts have been made to modify image emotions with color and style adjustments, facing limitations in effectively conveying emotions with fixed image contents.
	In this work, we introduce Emotional Image Content Generation (EICG), a new task to generate semantic-clear and emotion-faithful images given emotion categories.
	Specifically, we propose an emotion space and construct a mapping network to align it with the powerful Contrastive Language-Image Pre-training (CLIP) space, providing a concrete interpretation of abstract emotions.
	Attribute loss and emotion confidence are further proposed to ensure the semantic diversity and emotion fidelity of the generated images.
	Our method outperforms the state-of-the-art text-to-image approaches both quantitatively and qualitatively, where we derive three custom metrics, i.e., emotion accuracy, semantic clarity and semantic diversity.
	In addition to generation, our method can help emotion understanding and inspire emotional art design.
\vspace{-10pt}
\end{abstract}    
\section{Introduction}
\label{sec:intro}

\centerline{\textit{``What I cannot create, I do not understand.''}}
\rightline{\textit{--Richard Feynman}}

Emotions, often elusive yet profoundly influential, shape our actions, foster connections, and spark passions.
With the prevalence of social medias, users tend to share specially crafted images to express their feelings.
Aiming to find out people's emotional responses towards different stimuli, Visual Emotion Analysis (VEA) is an intriguing yet challenging task in computer vision~\cite{rao2016learning,yang2018weakly,yang2021solver}.
Recent years have witnessed rapid development in this field, bringing potential applications such as opinion mining~\cite{yadollahi2017current}, market advertising~\cite{consoli2010new} and mental healthcare~\cite{hsieh2015conceptualizing}.

Thanks to the advent of diffusion models~\cite{ho2020denoising,dhariwal2021diffusion,rombach2022high}, unprecedented progress has been made in text-to-image generation, where users can generate high-quality images with crafted prompts or personalized objects~\cite{gal2022image,ruiz2023dreambooth,zhang2023adding}.
Existing text-to-image diffusion models,  are often excel in generating \textit{concrete} concepts (\eg, \textit{cat}, \textit{house}, \textit{mountain}) but face limitations when tasked with more \textit{abstract} ones (\eg, \textit{amusement}, \textit{anger}, \textit{sadness}).
In reality, however, photographic works are not necessarily targeted on specific entities, but are often composed to convey certain feelings.

A natural question arises: 
\textit{What if machines could create images that not only please our eyes but also touch our hearts?}
Generating emotions is very challenging. 
Emotions are abstract while images are concrete, leaving the affective gap~\cite{hanjalic2006extracting} hard to surmount.
To bridge the gap, several efforts have been made to modify visual emotions by adjusting colors and styles, \ie, image emotion transfer~\cite{peng2015mixed,sun2023msnet,weng2023affective}.
These methods, however, meet difficulties in evoking emotions correctly and significantly, \ie, 29\% emotion accuracy~\cite{weng2023affective}, as fixed image contents limit emotional variations.
Moreover, we cannot generate emotional images solely from colors and styles. 
What truly triggers emotion?
Psychological studies show that visual emotions are often evoked by specific semantics~\cite{brosch2010perception,camras1980emotion,borth2013sentibank}.

\begin{figure}
	\centering
	\includegraphics[width=\linewidth]{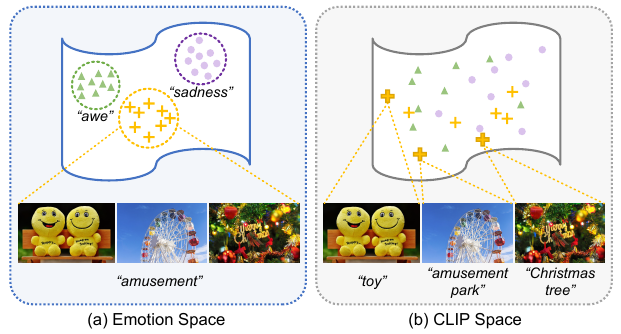}
	\vspace{-10pt}
	\caption{Despite (b) CLIP space demonstrates a powerful semantic structure, it struggles to effectively capture emotional relationships within (a), the proposed emotion space.}
	%	\Description{}
	\label{fig:emo_space}
	\vspace{-15pt}
\end{figure}

In this paper, we propose Emotional Image Content Generation (EICG), a new task to generate semantically clear and emotionally faithful visual contents conditioned on a given emotion category, as shown in~\Cref{fig:teaser}.
Semantic clarity demands an unambiguous representation of visual contents, while emotion faithfulness entails generating images evoke the intended emotions. 
Contrastive Language-Image Pre-training (CLIP)~\cite{radford2021learning} is a large-scale vision-language model with rich semantics.
However, we observe in~\Cref{fig:emo_space} that CLIP space can not well capture emotional relationships.
Therefore, we introduce an emotion space, which groups similar emotions together while keeping dissimilar ones apart.
While emotion space excels in representing emotions, CLIP space exhibits a powerful semantic structure.
To align emotion space with CLIP space, we propose a mapping network, interpreting abstract emotions with concrete semantics.

EmoSet~\cite{yang2023emoset} is a recently proposed large-scale visual emotion dataset with rich attributes.
The Latent Diffusion Model (LDM) loss~\cite{rombach2022high} is often utilized to optimize concrete entities with single and explicit semantics, posing a challenge in capturing the diversity within each emotion. 
To address this, we introduce an attribute loss to ensure semantic clarity and diversity, by leveraging the attribute labels in EmoSet.
Recognizing that not all objects are affective, emotion confidence is further proposed to ensure the emotion fidelity of the generated contents.

To estimate the generation quality of EICG, three evaluation metrics are specially designed: emotion accuracy, semantic clarity and semantic diversity.
As EICG aims to create emotional contents, we design emotion accuracy to measure the alignment between intended and perceived emotions in the generated images.
People are prone to evoke emotions only when the contents are easily recognizable.
Thus we propose semantic clarity to assess the unambiguity of the generated image content.
Additionally, in view of the assorted emotion stimuli, we devise semantic diversity to quantify the content richness under each emotion.
We evaluate our method through both qualitative and quantitative analyses, surpassing the state-of-the-art text-to-image generation approaches across five metrics.
Ablation studies are performed to verify the network design, and user studies are conducted to resonate our method with human viewers.
Besides generation task, our method can also be applied to decompose emotion concepts, transfer emotional contents and fuse different emotions, which may be helpful to understand emotions and create emotional art design.

In summary, our contributions are:
%\vspace{-0.4cm}
\begin{itemize}
	\setlength{\itemsep}{0pt}
	\setlength{\parsep}{0pt}
	\setlength{\parskip}{0pt}
	
	\item We introduce Emotional Image Content Generation, a novel task to generate emotion-faithful and semantic-clear image contents. We also derive three custom metrics to estimate the generation performance.
	
	\item We develop a mapping network to align the proposed emotion space to the powerful CLIP space, where attribute loss and emotion confidence are further designed to ensure the semantic richness and emotion fidelity.
	
	\item We evaluate our method against the state-of-the-art text-to-image approaches and demonstrate our superiority. Potential applications are exhibited for emotion understanding and emotional art design.
	
%	We verify that some specific semantics are deterministic in evoking visual emotions, where we visualize two settings under emotion-fixed varied semantics and neutral-semantic varied emotions. 

\end{itemize}

\section{Related work}
\label{sec:rw}

\subsection{Visual Emotion Analysis}

Researchers have been involved in VEA for over two decades, ranging from early traditional approaches~\cite{lee2011fuzzy,machajdik2010affective,borth2013large} to recent deep learning ones~\cite{rao2020learning,zhang2019exploring,yang2021stimuli,yang2021solver}.
Given the inherent abstractness and complexity of visual emotion, researchers aim to identify the most influential elements, which range from low-level features like color, texture and style~\cite{lee2011fuzzy,machajdik2010affective,rao2020learning,zhang2019exploring} to high-level semantics~\cite{borth2013large,rao2020learning,zhang2019exploring,yang2021stimuli,yang2021solver}.
Lee \etal~\cite{lee2011fuzzy} propose a scheme to evaluate emotional response from color images by reasoning the prototypical color for each emotion and the input images.
As a milestone, Machajdik \etal~\cite{machajdik2010affective} extract representative low-level features in composition, including color and texture, to predict visual emotions.
Besides low-level features, Borth \etal~\cite{borth2013large} propose Adjective-Noun Pair (ANP) and build a visual concept detector named Sentibank.
With the help of deep learning techniques, Rao \etal~\cite{rao2020learning} construct MldrNet to extract emotional clues from pixel-level, aesthetic-level and semantic-level.
To form a more discriminative emotional representation, Zhang~\etal~\cite{zhang2019exploring} integrate high-level contents and low-level styles.
Yang~\etal propose network to mine emotions from multiple objects~\cite{yang2021stimuli} as well as object-scene correlations~\cite{yang2021solver}.
Existing work often treat VEA as a classification task, \ie, input an image and predict the emotion within it.
Can we reverse this process? 
In other words, can we generate an image targeting on the given emotion word? 
Only by creating emotional images can we demonstrate the significance of visual elements, leading to a better understanding of emotions.

\subsection{Text-to-Image Generation}

Text-to-image generation aims to convert textual descriptions into corresponding realistic images.
Existing generative models can be grouped into GANs~\cite{goodfellow2020generative,liao2022text,zhu2019dm}, VAEs~\cite{kingma2013auto,gafni2022make,zhang2018stacking}, flow-based~\cite{rezende2015variational}, energy-based~\cite{lecun2006tutorial} and diffusion-based~\cite{ho2020denoising,rombach2022high,dhariwal2021diffusion,zhang2023adding,ruiz2023dreambooth}.
Diffusion models are witnessed impressive and rapid progress in recent years, where methods like GLIDE~\cite{nichol2021glide}, DALLE2~\cite{ramesh2022hierarchical}, Imagen~\cite{saharia2022photorealistic} can generate diverse, photo-realistic and high-quality images.
Notably, Stable diffusion~\cite{rombach2022high} is one of the most popular diffusion models, owing to its stable training and the capability for fine-grained control, supported by an active user community.
For customized generation, several diffusion-based text-to-image methods are introduced, where methods vary from learning a new embedding~\cite{gal2022image,dong2022dreamartist} and finetuning the network parameters~\cite{ruiz2023dreambooth,kumari2023multi,wei2023elite}.
Textual inversion~\cite{gal2022image} and DreamArtist~\cite{dong2022dreamartist} learn new concepts with a few user-provided images in the word embedding space, without further training on diffusion models.
While DreamBooth~\cite{ruiz2023dreambooth} finetunes all the parameters to learn a new concept, Custom diffusion~\cite{kumari2023multi} only updates the key and value parameters in the cross attention layers.
Further, ELITE~\cite{wei2023elite} speeds up the running time with accurate generation results by adopting a global and local mapping network.
Existing text-to-images models are capable of generating concrete concepts~\cite{dhariwal2021diffusion,zhang2023adding,liao2022text}, or personalized ones~\cite{gal2022image,ruiz2023dreambooth,kumari2023multi}, but face difficulties in generating more abstract ones.
In reality, photographic works are not necessarily composed of targeted concepts, but often aim to convey specific feelings.
Thus, how to generate emotion-evoking images remains a pressing and critical challenge.

\begin{figure*}
	\centering
	\includegraphics[width=0.88\linewidth]{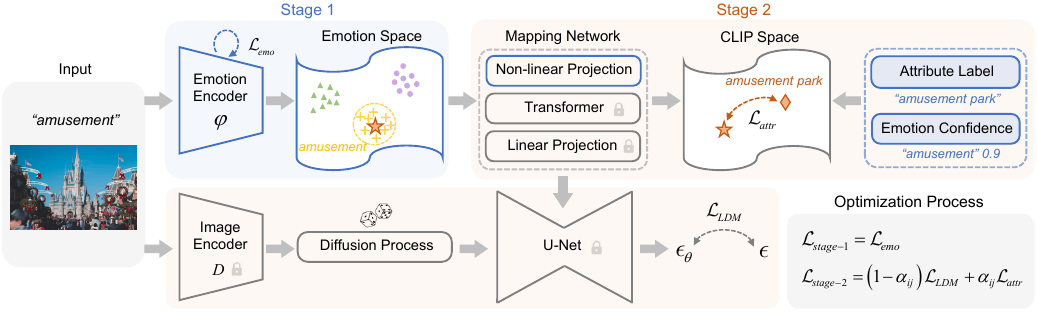}
	\vspace{-5pt}
	\caption{Training process of our network. Emotion representation (stage 1) learns a well-behaved emotion space and emotion content generation (stage 2) maps this space to CLIP space, aiming to generate image contents with emotion fidelity, semantic clarity and diversity.}
	%	\Description{}
	\label{fig:method-1}
	\vspace{-10pt}
\end{figure*}
\subsection{Image Emotion Transfer}

Image style transfer~\cite{gatys2016image} aims to render the semantic content under different styles, producing visually stunning results~\cite{karras2019style,wang2023stylediffusion,rangwani2023noisytwins,yang2022pastiche}.
Similarly, image color transfer~\cite{reinhard2001color} seeks to adjust and harmonize the color characteristics of one image to match another~\cite{huang2020learning,oskarsson2021robust}.
Specifically, color and style choices can strongly influence the emotions of an image~\cite{mohammad2018wikiart}.
By adjusting low-level visual elements, image emotion transfer aims to modify the emotional tone of the input image, including the color-based methods~\cite{yang2008automatic,wang2013affective,peng2015mixed,liu2018emotional,chen2020image,zhu2023emotional} and the style-based ones~\cite{sun2023msnet,weng2023affective}.
Yang and Peng~\etal~\cite{yang2008automatic} makes the first attempt to transfer image colors. 
Wang~\etal~\cite{wang2013affective} present a system to modify the image color according to a given emotion word, and Liu~\etal~\cite{liu2018emotional} further advance it with deep learning techniques.
Peng~\etal~\cite{peng2015mixed} introduce a new approach to alter the emotion of an input image by guiding its color and texture under the target image.
More recently, to reflect emotions in styles, Sun~\etal~\cite{sun2023msnet} and Weng~\etal~\cite{weng2023affective} bring promising results on emotion-aware image style transfer.
Nevertheless, the alteration of visual emotions through colors and styles is limited due to fixed content, resulting in subtle emotional changes, \ie, 29\% emotion accuracy in~\cite{weng2023affective}.
Psychological studies suggest that visual emotions can be elicited by specific semantics~\cite{brosch2010perception}.
Thus, we propose a novel method to generate emotional image contents with clear semantics.

\section{Method}
\label{sec:method}

%The training process of our method is shown in~\Cref{fig:method-1}, consisting of two stages, \ie, emotion representation and emotional content generation.
%We first learn a well-behaved emotion space and then map it to CLIP space to generate semantic-explicit and emotion-evoking contents. 

\subsection{Emotion Representation}

\paragraph{Emotion Space}
EICG is a challenging task, which requires both semantic clarity and emotion fidelity.
%On one hand, the generated image are expected to conform to human cognition, which requires semantic clarity.
%More importantly, our primary goal is to create emotion-evoking images, underscoring the importance of emotional consistency.
How to generate an image with distinct and emotional semantics?
CLIP~\cite{radford2021learning} is developed to align image and text modalities, where semantically related features are located in close proximity to each other.
While CLIP shows impressive semantic representation capabilities, it struggles to effectively capture emotional relationships.
As demonstrated in~\Cref{fig:emo_space}, we can observe that sample points with emotional similarities are distantly separated within the CLIP space due to their differing semantics, \eg, \textit{toy}, \textit{amusement park} and \textit{Christmas tree}.
To better depict emotional relationships, we introduce the emotion space, a latent space that clusters similar emotions together while keeping dissimilar ones apart.
EmoSet~\cite{yang2023emoset} is a large-scale dataset with rich attributes, where each image is labeled with an emotion.
Using aligned image-emotion pairs, we construct an encoder $\varphi$ with ResNet-50~\cite{he2016deep} to capture emotion representations.
%Naturally, we use the text-image pairs from EmoSet~\cite{yang2023emoset} to train an emotion encoder by aligning with different modalities.
%Particularly, we employ ResNet-50~\cite{he2016deep} and adopt the architecture before the final fully-connected layer. 
To train the encoder, we devise an emotion loss by implementing the widely-used Cross-Entropy (CE) loss, following the previous work~\cite{yang2018weakly,yang2021solver}:
\begin{align}
	\label{eq:l_emo}
	{{\mathcal{L}}_{emo}}=-\sum\limits_{i=1}^{C}{y_{emo}\log \frac{\exp (\varphi (x,i))}{\sum\nolimits_{i=1}^{C}{\exp (\varphi (x,i))}}},
\end{align}
where $x$ represents the input image, $y_{emo}$ denotes the emotion label and $C$ stands for the total number of emotion categories.
Once the loss function converges, emotion space is established.
Parameters in emotion encoder remain fixed in the following emotional content generation process.

During inference, each emotion cluster is represented by a Gaussian distribution with learned parameters, \ie, mean and standard deviation.
For example, when taking \textit{amusement} as input, we randomly sample a data point from corresponding Gaussian distribution to serve as its emotion representation, as shown in~\Cref{fig:method-1}.
We have confirmed that Gaussian distribution suits emotion clusters well and the random sampling process induces diversity to EICG.

\subsection{Emotional Content Generation}

\paragraph{Mapping Network}
While emotion space is emotionally separable, CLIP space captures rich semantics.
Existing text-to-image models entail clear and specific semantics as input, making CLIP space indispensable in the generation process.
Consequently, establishing the mapping between emotion space and CLIP space becomes a crucial challenge.
Intuitively, we attempt to build the mapping network using fully connected layers, following previous work~\cite{rangwani2023noisytwins,tevet2022motionclip}.

However, as depicted in Figure \ref{fig:emo_space}, clustered feature points in the emotion space are expected to disperse in the CLIP space to capture diverse semantics.
Therefore, we utilize a Multilayer Perceptron (MLP) to build the mapping network, incorporating non-linear operations, \ie, RELU, to facilitate the separation process.
The non-linear projection $F$ is succeeded by a CLIP text transformer $t_{\theta }$, yielding textual embedding for U-Net.
The end-token embedding of the transformer's output is passed through a fully-connected layer, producing the CLIP text feature.
Particularly, to better preserve the prior knowledge in the CLIP space, parameters in the transformer and linear projection are kept frozen, while parameters in non-linear projection are learned, as depicted in~\Cref{fig:method-1}.

\paragraph{Attribute Loss}
Existing text-to-image diffusion models often employ Latent Diffusion Model (LDM) loss~\cite{rombach2022high} for optimization process~\cite{gal2022image,ruiz2023dreambooth,zhang2023adding}:
\begin{align}
	\label{eq:l_ldm}
	{{\mathcal{L}}_{LDM}}={{\mathbb{E}}_{z,x,\epsilon,t}}\!\left[ \left\| \epsilon \!-\!{{\epsilon }_{\theta }}\left( {{z}_{t}},t,{t_{\theta }}\left( {{F}}\left( \varphi \left( x \right) \right) \right) \right) \right\|_{2}^{2} \right],
\end{align}
%\begin{align}
%	\label{eq:l_ldm_1}
%	y={{F}_{n}}\left( \varphi \left( x \right) \right),
%\end{align}
where $\epsilon$ represents the added noise, ${\epsilon }_{\theta }$ denotes the denoising network and $z_t$ indicates the latent noised to time $t$.

\begin{figure}
	\centering
	\includegraphics[width=0.9\linewidth]{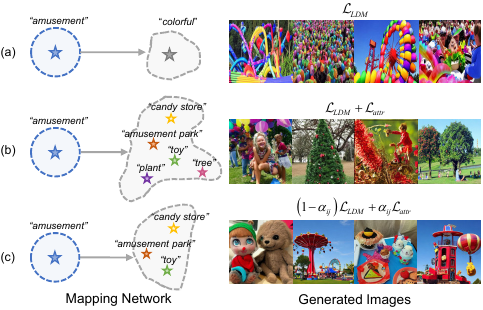}
	\vspace{-5pt}
	\caption{Motivation for loss function design. Compare to (a) LDM loss alone, (b) attribute loss enhances semantic clarity while (c) emotion confidence ensures emotion accuracy.}
	%	\Description{}
	\label{fig:method-2}
	\vspace{-10pt}
\end{figure}

In these cases, target concepts typically involve concrete entities (\eg, \textit{dog}, \textit{car}, \textit{flower}) or personalized objects (\eg, someone's \textit{corgi}).
These concepts often exhibit consistency on semantic level and share certain similarities on pixel level.
However, emotions are abstract concepts, where multiple semantics coexist under one specific category.
Learning emotions solely with LDM loss may pose some challenges.
For one thing, each emotion might collapse to a specific semantic point, \eg, \textit{amusement} collapsing to \textit{amusement park}, losing intra-class diversity.
In reality, semantics within one emotion are diverse, where single point cannot fully capture.
Moreover, since LDM loss is designed to reconstruct the input image, it primarily focuses on learning and preserving pixel-level commonalities such as color and texture. 
In~\Cref{fig:method-2} (a), with LDM loss alone, CLIP embedding for \textit{amusement} is prone to be \textit{colorful}, without exhibiting explicit and diverse semantics.
We can conclude that it is hard to achieve robust emotion representations in CLIP space by implementing LDM loss alone.

In the pursuit of clear and diverse contents, semantics guidance is essential for the generation process.
Thanks to the rich attribute annotations in EmoSet, we select the mid-level attributes, \ie, object class and scene type to guide the generation process.
With this semantic guidance, we formulate an attribute loss to guarantee that the generated image contents possess clear and diverse semantics.
For clarity, emotions are easily triggered in people only when visual contents are represented in an unambiguous manner.
Considering the varied emotional stimuli in reality, attribute loss guides the network to learn multiple semantics under one specific emotion.
%For clarity, only when contents are explicit can human understand image well, leading to stronger emotional arousal.
%The one-to-many emotion-semantic correspondences conform to real-world scenarios, which is also consistent with human cognition.
Our attribute loss is devised on CLIP space, by calculating the cosine similarities $f(\cdot)$ and optimizing a symmetric CE loss over the similarity scores~\cite{radford2021learning}:
\begin{align}
	\label{eq:l_attr}
	{{\mathcal{L}}_{attr}}\!=\!-\!\sum\limits_{j=1}^{K}{{{y}_{attr}}\log \frac{\exp \!\left( f\!\left( {{v}_{emo}},\!{{\tau }_{\theta }}\!\left({a}_{j} \right) \right) \right)}{\sum\nolimits_{j=1}^{K}\!{\exp \!\left( f\!\left( {{v}_{emo}},\!{{\tau }_{\theta }}\!\left( {a}_{j} \right) \right) \right)}}},
\end{align}
\begin{align}
	\label{eq:cos}
	f\left( p,q \right)=\frac{p\cdot q}{\left\| p \right\|\left\| q \right\|},
\end{align}
where $a_j$ denotes the $j$ member in the attribute set, ${{\tau }_{\theta }}$ represents the text encoder, $v_{emo}$ implies the learned CLIP embedding and $K$ indicates the total number of the attributes.
With the attribute loss, each sample point is converging towards the correct semantic and distancing itself from the incorrect ones.
Through the combination of attribute loss and LDM loss, we can effectively map each emotion to clear and diverse semantics, as demonstrated in~\Cref{fig:method-2} (b).

\begin{figure}
	\centering
	\includegraphics[width=0.9\linewidth]{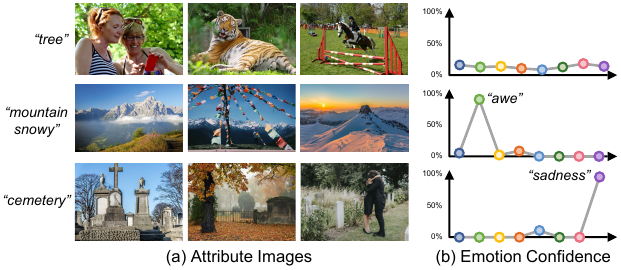}
	\vspace{-5pt}
	\caption{Illustration of emotion confidence. Each (a) attribute is represented by (b) a distribution of confidence on eight emotions.}
	%	\Description{}
	\label{fig:emo_confidence}
	\vspace{-10pt}
\end{figure}

\paragraph{Emotion Confidence}
However, it is worth noting that some of the semantics in~\Cref{fig:method-2} (b) appear emotionally neutral, \eg, \textit{plant} and \textit{tree}.
Since attributes are annotated objectively, not all the attributes in EmoSet are emotional.
Therefore, we propose emotion confidence to measure the correlations between emotions and semantic attributes.
Initially, we gather all images associated with attribute $j$ in EmoSet and send them to a pre-trained emotion classifier.
Each image is predicted as an emotion vector ${p}(\cdot)$ and we sum all images up to get the emotional distribution $d_j$ for attribute $j$. Each emotion $i$ within this distribution is assigned a corresponding emotion confidence ${\alpha }_{ij}$:
\begin{align}
	\label{eq:emo_conf}
	{{\alpha }_{ij}}=\frac{1}{{{N}_{j}}}\sum\limits_{n=1}^{{{N}_{j}}}{{{p}}\left( {{x}_{n}}, i \right)},
\end{align}
where $x_n$ represents the input image and $N_j$ denotes the total image number in attribute $j$.
We further illustrate the above process in~\Cref{fig:emo_confidence} with visual representations.
When \textit{mountain snowy} appears, people are more likely to experience \textit{awe} and \textit{cemetery} often elicits \textit{sadness}.
%These semantics have close relationships towards one specific emotion. 
In contrast, the presence of \textit{tree} in every emotion category suggests its lack of emotional specificity. 
Some attributes are emotion-related while others are not, which can be beneficial for generating emotional contents.
We then use emotion confidence to balance between LDM loss and attribute loss:
\begin{align}
	\label{eq:loss}
	{\mathcal{L}}=\left( 1-{\alpha}_{ij}  \right){{\mathcal{L}}_{LDM}}+{\alpha}_{ij} {{\mathcal{L}}_{attr}},
\end{align}
where $i$ represents the emotion category $y_{emo}$ and $j$ denotes the attribute type $y_{att}$.
The greater the emotion confidence ${\alpha}_{ij}$ is, the stronger the impact attribute $j$ has on the specific emotion $i$.
Low confidence suggests a weak connection between the attribute and emotion, signaling that the network should learn more from the pixel-wise LDM loss.
When higher confidence occurs, the network should prioritize the semantic meaning of the image, \ie, the attribute loss.
%With such design, our network can adaptive to different cases to learn visual emotions, resulting in generating both semantic explicit and emotion consistent image contents.
With this design, our network can adapt to a wide range of cases, generating image contents that are both semantically explicit and emotionally faithful, as shown in~\Cref{fig:method-2} (c).

\begin{figure*}
	\centering
	\includegraphics[width=\linewidth]{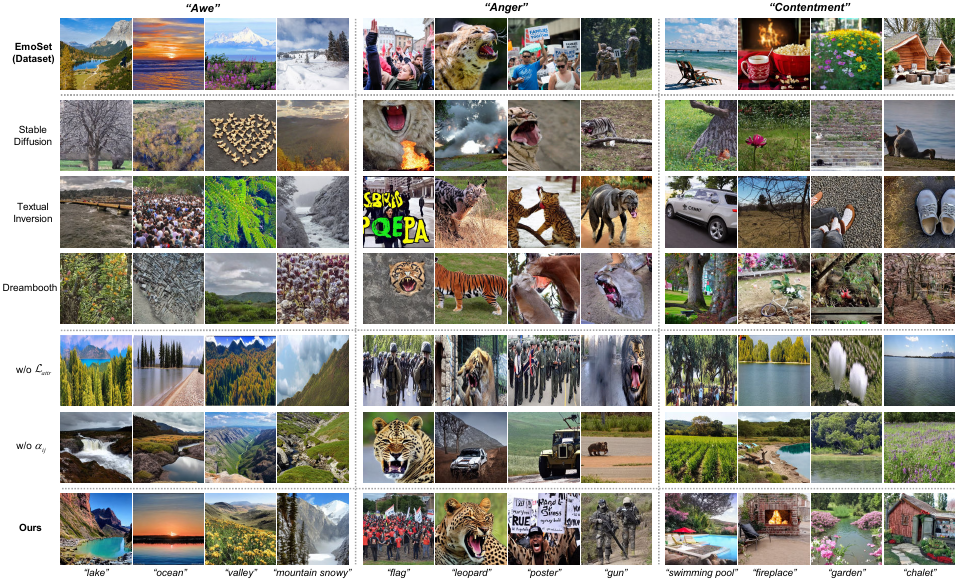}
	\vspace{-15pt}
	\caption{Qualitative comparisons with the state-of-the-art text-to-image generation approaches and ablation studies of our method.}
	%	\Description{}
	\label{fig:exp_1}
	\vspace{-15pt}
\end{figure*}

%\subsection{Inference}
\section{Experiments}
\label{sec:exp}
\subsection{Dataset and Evaluation}

\paragraph{Dataset}

EmoSet~\cite{yang2023emoset} is a large-scale visual emotion dataset with rich attributes, comprising a total of 118,102 images.
To investigate the connections between emotions and specific contents, we create a subset from EmoSet by preserving images with object/scene labels.
Each image is labeled with both emotion and attribute labels, guiding the optimization process of emotion loss and attribute loss.
Notably, the wide range of attribute labels assures for learning diverse and representative emotional contents.

\paragraph{Evaluation Metrics}
To comprehensively evaluate the performance of different methods on EICG task, we utilize commonly used metrics (FID, LPIPS) and design some specific ones (Emo-A, Sem-C, Sem-D).
1) \textbf{FID}: Frechet Inception Distance (FID)~\cite{heusel2017gans} quantifies the distribution distance between generated and real images, providing an estimate of image fidelity.
2) \textbf{LPIPS}: Similar to~\cite{tan2021diverse}, we employ LPIPS~\cite{zhang2018unreasonable} to assess the overall image diversity, with higher values indicating better performance. 
3) \textbf{Emo-A}: Since EICG aims at creating emotion-evoking images, we design emotion accuracy to assess the emotional alignment between the targeted emotions and the generated images.
4) \textbf{Sem-C}: People are easily to evoke emotions under recognizable contents. 
We thus introduce semantic clarity to assess the explicitness of generated image contents.
5) \textbf{Sem-D}: Emotions are complex, where each can be triggered by multiple factors.
To cover a diverse range of potential scenes or objects, we derive semantic diversity to estimate the content richness associated with each emotion.
For more details, please refer to the supplementary materials.

\subsection{Comparisons}
\begin{table}
	\centering
%		\normalsize
	\scriptsize
	\caption{Comparisons with the state-of-the-art methods and ablation studies on emotion generation task, involving five metrics.}
	\vspace{-5pt}
	\label{tab:exp_sota}
	\renewcommand\arraystretch{1.18}
	\setlength\tabcolsep{4pt}
	\begin{tabular}{lcccccccc}
		\toprule
		Method & FID $\downarrow$ & LPIPS $\uparrow$ & Emo-A $\uparrow$ & Sem-C $\uparrow$ & Sem-D $\uparrow$  \\
		%		\midrule
		%		AlexNet~\cite{krizhevsky2012imagenet} &&&&&&&& \\
		%		VGG-16~\cite{simonyan2014very} &&&&&&&& \\
		%		ResNet-50~\cite{he2016deep} &&&&&&&& \\
		%		CLIP~\cite{radford2021learning} &&&&&&&& \\
		\midrule
		Stable Diffusion~\cite{rombach2022high} & 44.05 & 0.687 & 70.77\% & 0.608 & 0.0199  \\
		Textual Inversion~\cite{gal2022image} & 50.51 & 0.702 & 74.87\% & 0.605 & 0.0282  \\
		DreamBooth~\cite{ruiz2023dreambooth}  & 46.89 & 0.661 & 70.50\% & 0.614 &  0.0178 \\
%		\rowcolor{mygray} 
		Ours & \textbf{41.60} & \textbf{0.717} & \textbf{76.25\%} &\textbf{0.633} &  \textbf{0.0335} \\
%		CLIP~\cite{radford2021learning} & 86.96 & 86.44 & 88.03 & 88.61 & 71.21 & 78.70 &  & \textbf{80.00}\\
		\midrule
%		w/o emotion space &  &  &  &  &  \\
		w/o $F$  & 57.54 & 0.713 & 71.12\% & 0.615 & 0.0261 \\
		w/o ${\mathcal{L}}_{attr}$ & 51.13 & 0.707 & 65.75\% & 0.592 & 0.0270 \\
		w/o ${{\alpha}}_{ij}$ & 43.30 & 0.714 & 74.88\% & 0.591 & 0.0263 \\
%		SOLVER~\cite{yang2021solver}  & 85.43 & 83.19 & 86.20 & 85.60 & 62.12 & 72.33 & & \\
		\bottomrule
	\end{tabular}
\vspace{-15pt}
\end{table}

As our method is the first attempt in EICG, we compare it with the most relevant and state-of-the-art text-to-image generation techniques: Stable diffusion~\cite{rombach2022high}, Textual inversion~\cite{gal2022image} and Dreambooth~\cite{ruiz2023dreambooth}.
While Stable diffusion is a general image generation pipeline, Textual inversion and Dreambooth specialize in customized image generation.

\paragraph{Qualitative Comparisons}
In~\Cref{fig:exp_1}, our method is qualitatively compared with the state-of-the-art methods across three emotion categories, \ie, \textit{awe}, \textit{anger} and \textit{contentment}.
Generation results of the rest five emotions can be found in the supplementary materials.
Take \textit{awe} as an example, all the three compared methods tend to produce images with dense textures and dim colors, which suggests that representations for each emotion may collapse to a single feature point.
%Emotions are complex and thus are hard to be represented by a single feature point.
For \textit{anger} and \textit{contentment}, both Stable diffusion and Dreambooth distort the visual representations, \eg, \textit{tiger} and \textit{bicycle}, and generate some contents with ambiguous semantics.
Though Textual inversion preserves some semantic fidelity, it generates emotion-agnostic contents such as \textit{shoes} and \textit{cars}.
Since these methods are crafted to learn customized concepts, challenges may arise when handling complex and diverse emotional images.
Rather than generating \textit{plants} and \textit{trees}, our method can provide diverse and emotion-evoking image contents for \textit{awe} through \textit{lakes}, \textit{oceans}, \textit{valleys} and \textit{snow-covered mountains}.
In \textit{anger}, our approach extends beyond mere \textit{beasts}, encompassing \textit{flags}, \textit{posters}, and \textit{guns}.
Owing to attribute loss and emotion confidence, our method can effectively capture the rich and varied semantics while maintaining emotion faithfulness in EmoSet.

\begin{table}
	\centering
%		\normalsize
	\scriptsize
	\caption{User preference study. The numbers indicate the percentage of participants who prefer our results over those compared methods, given the same emotion category as input.}
	\vspace{-5pt}
	\label{tab:exp_userstudy}
	\renewcommand\arraystretch{1.18}
	\setlength\tabcolsep{2.5pt}
	\begin{tabular}{lcccccccc}
		\toprule
		Method & Image fidelity $\uparrow$ & Emotion faithfulness $\uparrow$ & Semantic diversity $\uparrow$ \\
		%		\midrule
		%		AlexNet~\cite{krizhevsky2012imagenet} &&&&&&&& \\
		%		VGG-16~\cite{simonyan2014very} &&&&&&&& \\
		%		ResNet-50~\cite{he2016deep} &&&&&&&& \\
		%		CLIP~\cite{radford2021learning} &&&&&&&& \\
		\midrule
		Stable Diffusion & 67.86$\pm$15.08\% & 73.66$\pm$11.80\% & 87.88$\pm$9.64\%  \\
		Textual Inversion & 79.91$\pm$16.92\%& 72.75$\pm$16.90\% & 85.66$\pm$10.51\%   \\
		DreamBooth & 77.23$\pm$14.00\%& 80.79$\pm$8.64\% & 81.68$\pm$17.06\%  \\
%		SOLVER~\cite{yang2021solver}  & 85.43 & 83.19 & 86.20 & 85.60 & 62.12 & 72.33 & & \\
		\bottomrule
	\end{tabular}
\vspace{-15pt}
\end{table}

\paragraph{Quantitative Comparisons}
As shown in~\Cref{tab:exp_sota}, the proposed method surpasses the compared methods across all five evaluation metrics.
Particularly, better performance on FID and LPIPS indicates our method can generate images with higher fidelity and diversity, effectively capturing the characteristics of the training data.
All methods achieve comparable results on emotion accuracy.
From~\Cref{fig:exp_1}, we observe that comparison methods are prone to fall into singular or incorrect emotion representations.
Even such generation results are still separable in eight classes, they do not conform to human emotional cognition.
%Although each emotion may collapse to a single representation, it can still exhibit high emotional accuracy.
This suggests that relying solely on Emo-A may be insufficient for EICG task.
Therefore, we additionally introduce Sem-C and Sem-D to estimate the content clarity and diversity, where our method exhibits a clear advantage over other methods.

\paragraph{User Study}
Besides qualitative and quantitative comparisons, we also conduct a user study to determine whether our method is preferred by humans and to understand how people perceive emotions.
We invite 14 participants from different social backgrounds and each test session lasts about 30 minutes.
In the first part, generation results are evaluated on three dimensions: image fidelity, emotion faithfulness and semantic diversity.
%, where comparison methods include Stable diffusion, Textual inversion and Dreambooth.
Each question presented to the participants includes two sets of images conveying the same emotion, drawn from our method and one of the comparison methods.
The participants are then asked: \textit{which group is more realistic? which group evokes a stronger sense of [emotion type]? which group is more diverse?}
As illustrated in~\Cref{tab:exp_userstudy}, our method attains the top rankings compared to the other three methods, particularly excelling in semantic diversity.
We aim to explore the factors influencing visual emotions in the second part.
Participants are shown an emotional image generated by our method and are asked: \textit{which emotion best describes the image? why do you feel such emotion?}
Compared to the 76.25\% machine predicted one in~\Cref{tab:exp_sota}, 82.14\% emotion accuracy is achieved by user voting, where generated images are more emotion-evoking towards human participants. 
Additionally, 88.39\% of the responses indicate that emotions are predominantly triggered by the content/semantic.
This underscores how our task, EICG, is closely aligned with human cognition.
%Specifically, emotion faithfulness shows whether the generated images are emotional or not and semantic diversity reveals whether the generated emotional contents diverse or not.

\begin{figure}
	\centering
	\includegraphics[width=\linewidth]{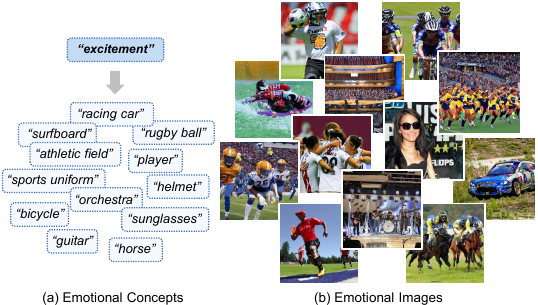}
	\vspace{-5pt}
	\caption{Emotion decomposition. Each emotion word is broken down into a set of (a) emotional concepts, reflecting the semantics in (b) generated images.}
	%	\Description{}
	\label{fig:exp_3}
	\vspace{-10pt}
\end{figure}
\begin{table}
	\centering
	% \normalsize
	% \footnotesize
	\scriptsize
	\caption{Comparisons with the state-of-the-art methods on emotion transfer task, involving three metrics. }
	\vspace{-5pt}
	\label{tab:exp_obj}
	\renewcommand\arraystretch{1.18}
	\setlength\tabcolsep{2.6pt}
	\begin{tabular}{lcccccc}
		\toprule
		\multirow{2}{*}{Method} & \multicolumn{2}{c}{Emo-A $\uparrow$} & \multicolumn{2}{c}{CLIP-img $\uparrow$} & \multicolumn{2}{c}{CLIP-txt $\uparrow$} \\
		%		\cmidrule(r){2-4} \cmidrule(r){5-7}
		& \textit{amusement} & \textit{fear} & \textit{amusement} & \textit{fear} & \textit{amusement} & \textit{fear} \\
		%		\multirow{2}{*}{Backbone} & \multicolumn{3}{c}{w/o pretrained} & \multicolumn{3}{c}{w/ pretrained} \\
		%		& w/o attr & w/ attr & Diff. & w/o attr & w/ attr & Diff. \\
		\midrule
		Stable Diffusion & 51.54\% & 56.67\% & \textbf{0.929} & 0.825 & 0.257 & 0.251\\
		Textual Inversion & 60.82\% & 40.00\% & 0.902 & 0.792 & 0.270 & 0.259\\
		Ours & \textbf{72.16\%} & \textbf{63.33\%} & 0.913 & \textbf{0.841} & \textbf{0.276} & \textbf{0.270}\\
		%		\midrule
		%		Fine-tuned AlexNet &&\\
		%		Fine-tuned VGG-16 &&\\
		%		Fine-tuned ResNet-50 &&\\
		%		Fine-tuned DensNet \\
		%		Fine-tuned CLIP &&\\
		\bottomrule
	\end{tabular}
	\vspace{-5pt}
\end{table}

\begin{figure*}
	\centering
	\includegraphics[width=\linewidth]{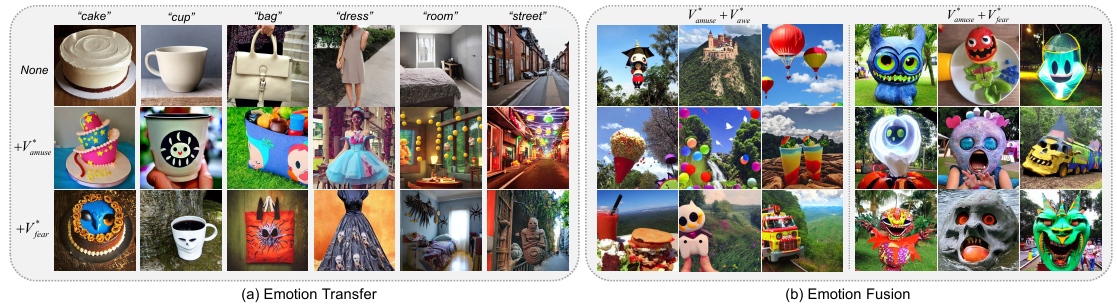}
	\vspace{-15pt}
	\caption{Emotion creation. (a) transfers emotion representations (\ie, \textit{amusement}, \textit{fear}) to a series of neutral contents while (b) fuse two emotions (\ie, \textit{amusement-awe}, \textit{amusement-fear}) together, which may be helpful for emotional art design.}
	%	\Description{}
	\label{fig:exp_5}
	\vspace{-10pt}
\end{figure*}

%\subsection{Emotion-Semantic Analysis}
\subsection{Ablation Study}
We examine the efficacy of each network design, encompassing the non-linear mapping network $F$, the attribute loss ${\mathcal{L}}_{attr}$ and the emotion confidence ${\alpha}_{ij}$.
In~\Cref{tab:exp_sota}, without nonlinear mapping network, emotion representations are aggregated, which fails to restore the real image set (high FID) and lacks semantic diversity (low Sem-D).
Attribute loss is introduced to enhance semantic clarity and diversity, whose absence leads to performance drops in Sem-C and Sem-D.
Besides, as shown in~\Cref{fig:exp_1}, generated images exhibit semantic distortions when attribute loss is absent (w/o ${\mathcal{L}}_{attr}$) and display explicit contents with attribute loss (w/o ${\alpha}_{ij}$).
While image contents become clear and diverse with attribute loss, it is only with emotion confidence that we can effectively filter out emotion-agnostic semantics and generate images that evoke specific emotions (Ours).

\subsection{Applications}
%Our method aims at generating emotional image content.
%On one hand, each emotion can be decomposed to various semantic contents, as shown in~\Cref{fig:exp_3}.
%On the other hand, these contents are decisive to evoke corresponding emotions, aiding emotional design in~\Cref{fig:exp_5}.

\paragraph{Emotion Decomposition}
Emotions, serving as abstract concepts, pose a challenge for generative models to understand.
Our method provides an opportunity to comprehend visual emotions by identifying the most relevant semantic contents for each emotion.
To be specific, we visualize the semantics that are most closely aligned with our emotion representations in CLIP space.
Each concept in~\Cref{fig:exp_3} (a), such as \textit{surfboard}, \textit{bicycle} and \textit{athletic field}, is very likely to elicit \textit{excitement}, where the corresponding images are presented in~\Cref{fig:exp_3} (b).
Upon viewing such images, we identify the semantics and instinctively link them to specific emotions.
These emotional concepts exhibit diversity, explicitness, and a strong capacity to evoke emotions.
By decomposing visual emotions, we can not only generate emotional images with various semantics but also gain a deeper understanding of emotion evocation process.
The results reveal the close relationship between emotions and semantics, in accordance with the psychological studies~\cite{brosch2010perception}.

\paragraph{Emotion Transfer}

Once we identify emotional contents, the next step is to explore how we can use it to create meaningful and compelling designs.
In addition to emotional content, there are also neutral ones.
As shown in~\Cref{fig:exp_5} (a), we combine the common neutral objects/scenes with emotional representations learned by our method.
Surprisingly, we find that these representations effectively preserve emotional semantics and seamlessly integrate them with new concepts.
Taking \textit{amusement} as an example, it preserves several semantics including \textit{amusement park}, \textit{picnic}, \textit{princess}, \textit{balloon} and \textit{beautiful lanterns}.
In~\Cref{tab:exp_obj}, our method is quantitatively compared with the state-of-the-art methods on emotion transfer task, specializing in \textit{room}, where our method can well-preserve semantics and effectively elicit emotions.   
Crucially, these creations can evoke explicit and strong emotions across various neutral semantics, suggesting the potential of our method in image editing, image transfer and emotional art design.

\paragraph{Emotion Fusion}

Additionally, we explore the possibilities of combining different emotion representations to evoke multiple emotions.
In \Cref{fig:exp_5} (b), we combine \textit{amusement} and \textit{awe} (positive-positive) as well as \textit{amusement} and \textit{fear} (positive-negative), bringing some intriguing observations.
In the combination of \textit{amusement} and \textit{awe}, we observe elements associated with \textit{amusement}, such as \textit{toys}, \textit{balloons}, and \textit{ice-creams}, alongside awe-inspiring elements like the blue \textit{sky}, \textit{mountains}, \textit{ocean}, and \textit{city views}.
Particularly, one may feel both fear and amusement when viewing the funny and horrible face.
When we fuse emotions, we are essentially combining their corresponding visual contents.

\section{Conclusion}
\label{sec:conclusion}

\paragraph{Discussion}
In this paper, we introduce a new task named EICG and derive three specially designed metrics.
We propose an emotion space and align it with the CLIP space, incorporating attribute loss and emotion confidence to ensure semantic clarity, semantic diversity and emotion fidelity.
Experimental results indicate that our method surpasses the state-of-the-art text-to-image diffusion models both qualitatively and quantitatively, where user study confirms its superiority.
Additionally, we outline potential applications for EICG and present some initial but promising results.

\paragraph{Limitations}
Emotions can be evoked by various visual factors such as color, style and content.
In this paper, we focus on investigating the most influential factor, \ie, contents.
Generating emotional images could be enhanced by considering a broader range of visual elements beyond content alone.
This is an avenue we plan to explore in future research.
Moreover, the relationships between emotions and content is not strictly binary.
In this paper, we simplify this connection by assuming content to be either emotional or emotion-agnostic. 
However, in reality, it is hard to assign \textit{rose} to a single emotion category.
\textit{White rose} may evoke \textit{sadness} while \textit{red rose} can elicit \textit{amusement}, making it hard to decide whether \textit{rose} is emotional or not.

{
	\small
	\bibliographystyle{ieeenat_fullname}
	\bibliography{EmoGen}

\begin{thebibliography}{60}
\providecommand{\natexlab}[1]{#1}
\providecommand{\url}[1]{\texttt{#1}}
\expandafter\ifx\csname urlstyle\endcsname\relax
  \providecommand{\doi}[1]{doi: #1}\else
  \providecommand{\doi}{doi: \begingroup \urlstyle{rm}\Url}\fi

\bibitem[Borth et~al.(2013{\natexlab{a}})Borth, Chen, Ji, and
  Chang]{borth2013sentibank}
Damian Borth, Tao Chen, Rongrong Ji, and Shih-Fu Chang.
\newblock Sentibank: large-scale ontology and classifiers for detecting
  sentiment and emotions in visual content.
\newblock In \emph{Proceedings of the 21st ACM International Conference on
  Multimedia}, pages 459--460, 2013{\natexlab{a}}.

\bibitem[Borth et~al.(2013{\natexlab{b}})Borth, Ji, Chen, Breuel, and
  Chang]{borth2013large}
Damian Borth, Rongrong Ji, Tao Chen, Thomas Breuel, and Shih-Fu Chang.
\newblock Large-scale visual sentiment ontology and detectors using adjective
  noun pairs.
\newblock In \emph{Proceedings of the 21st ACM International Conference on
  Multimedia}, pages 223--232, 2013{\natexlab{b}}.

\bibitem[Brosch et~al.(2010)Brosch, Pourtois, and Sander]{brosch2010perception}
Tobias Brosch, Gilles Pourtois, and David Sander.
\newblock The perception and categorisation of emotional stimuli: A review.
\newblock \emph{Cognition and Emotion}, 24\penalty0 (3):\penalty0 377--400,
  2010.

\bibitem[Camras(1980)]{camras1980emotion}
Linda Camras.
\newblock Emotion: a psychoevolutionary synthesis, 1980.

\bibitem[Chen et~al.(2020)Chen, Xiong, Zheng, and Luo]{chen2020image}
Tianlang Chen, Wei Xiong, Haitian Zheng, and Jiebo Luo.
\newblock Image sentiment transfer.
\newblock In \emph{Proceedings of the 28th ACM International Conference on
  Multimedia}, pages 4407--4415, 2020.

\bibitem[Consoli(2010)]{consoli2010new}
Domenico Consoli.
\newblock A new concept of marketing: The emotional marketing.
\newblock \emph{BRAND. Broad Research in Accounting, Negotiation, and
  Distribution}, 1\penalty0 (1):\penalty0 52--59, 2010.

\bibitem[Dhariwal and Nichol(2021)]{dhariwal2021diffusion}
Prafulla Dhariwal and Alexander Nichol.
\newblock Diffusion models beat gans on image synthesis.
\newblock \emph{Advances in Neural Information Processing Systems},
  34:\penalty0 8780--8794, 2021.

\bibitem[Dong et~al.(2022)Dong, Wei, and Lin]{dong2022dreamartist}
Ziyi Dong, Pengxu Wei, and Liang Lin.
\newblock Dreamartist: Towards controllable one-shot text-to-image generation
  via contrastive prompt-tuning.
\newblock \emph{arXiv preprint arXiv:2211.11337}, 2022.

\bibitem[Gafni et~al.(2022)Gafni, Polyak, Ashual, Sheynin, Parikh, and
  Taigman]{gafni2022make}
Oran Gafni, Adam Polyak, Oron Ashual, Shelly Sheynin, Devi Parikh, and Yaniv
  Taigman.
\newblock Make-a-scene: Scene-based text-to-image generation with human priors.
\newblock In \emph{European Conference on Computer Vision}, pages 89--106.
  Springer, 2022.

\bibitem[Gal et~al.(2022)Gal, Alaluf, Atzmon, Patashnik, Bermano, Chechik, and
  Cohen-Or]{gal2022image}
Rinon Gal, Yuval Alaluf, Yuval Atzmon, Or Patashnik, Amit~H Bermano, Gal
  Chechik, and Daniel Cohen-Or.
\newblock An image is worth one word: Personalizing text-to-image generation
  using textual inversion.
\newblock \emph{arXiv preprint arXiv:2208.01618}, 2022.

\bibitem[Gatys et~al.(2016)Gatys, Ecker, and Bethge]{gatys2016image}
Leon~A Gatys, Alexander~S Ecker, and Matthias Bethge.
\newblock Image style transfer using convolutional neural networks.
\newblock In \emph{Proceedings of the IEEE/CVF Conference on Computer Vision
  and Pattern Recognition}, pages 2414--2423, 2016.

\bibitem[Goodfellow et~al.(2020)Goodfellow, Pouget-Abadie, Mirza, Xu,
  Warde-Farley, Ozair, Courville, and Bengio]{goodfellow2020generative}
Ian Goodfellow, Jean Pouget-Abadie, Mehdi Mirza, Bing Xu, David Warde-Farley,
  Sherjil Ozair, Aaron Courville, and Yoshua Bengio.
\newblock Generative adversarial networks.
\newblock \emph{Communications of the ACM}, 63\penalty0 (11):\penalty0
  139--144, 2020.

\bibitem[Hanjalic(2006)]{hanjalic2006extracting}
Alan Hanjalic.
\newblock Extracting moods from pictures and sounds: Towards truly personalized
  tv.
\newblock \emph{IEEE Signal Processing Magazine}, 23\penalty0 (2):\penalty0
  90--100, 2006.

\bibitem[He et~al.(2016)He, Zhang, Ren, and Sun]{he2016deep}
Kaiming He, Xiangyu Zhang, Shaoqing Ren, and Jian Sun.
\newblock Deep residual learning for image recognition.
\newblock In \emph{Proceedings of the IEEE/CVF Conference on Computer Vision
  and Pattern Recognition}, pages 770--778, 2016.

\bibitem[Heusel et~al.(2017)Heusel, Ramsauer, Unterthiner, Nessler, and
  Hochreiter]{heusel2017gans}
Martin Heusel, Hubert Ramsauer, Thomas Unterthiner, Bernhard Nessler, and Sepp
  Hochreiter.
\newblock Gans trained by a two time-scale update rule converge to a local nash
  equilibrium.
\newblock \emph{Advances in Neural Information Processing Systems}, 30, 2017.

\bibitem[Ho et~al.(2020)Ho, Jain, and Abbeel]{ho2020denoising}
Jonathan Ho, Ajay Jain, and Pieter Abbeel.
\newblock Denoising diffusion probabilistic models.
\newblock \emph{Advances in Neural Information Processing Systems},
  33:\penalty0 6840--6851, 2020.

\bibitem[Hsieh and Nicodemus(2015)]{hsieh2015conceptualizing}
Elaine Hsieh and Brenda Nicodemus.
\newblock Conceptualizing emotion in healthcare interpreting: A normative
  approach to interpreters’ emotion work.
\newblock \emph{Patient Education and Counseling}, 98\penalty0 (12):\penalty0
  1474--1481, 2015.

\bibitem[Huang et~al.(2020)Huang, Qiu, Wang, and Li]{huang2020learning}
Yifei Huang, Sheng Qiu, Changbo Wang, and Chenhui Li.
\newblock Learning representations for high-dynamic-range image color transfer
  in a self-supervised way.
\newblock \emph{IEEE Transactions on Multimedia}, 23:\penalty0 176--188, 2020.

\bibitem[Karras et~al.(2019)Karras, Laine, and Aila]{karras2019style}
Tero Karras, Samuli Laine, and Timo Aila.
\newblock A style-based generator architecture for generative adversarial
  networks.
\newblock In \emph{Proceedings of the IEEE/CVF Conference on Computer Vision
  and Pattern Recognition}, pages 4401--4410, 2019.

\bibitem[Kingma and Welling(2013)]{kingma2013auto}
Diederik~P Kingma and Max Welling.
\newblock Auto-encoding variational bayes.
\newblock \emph{arXiv preprint arXiv:1312.6114}, 2013.

\bibitem[Kumari et~al.(2023)Kumari, Zhang, Zhang, Shechtman, and
  Zhu]{kumari2023multi}
Nupur Kumari, Bingliang Zhang, Richard Zhang, Eli Shechtman, and Jun-Yan Zhu.
\newblock Multi-concept customization of text-to-image diffusion.
\newblock In \emph{Proceedings of the IEEE/CVF Conference on Computer Vision
  and Pattern Recognition}, pages 1931--1941, 2023.

\bibitem[LeCun et~al.(2006)LeCun, Chopra, Hadsell, Ranzato, and
  Huang]{lecun2006tutorial}
Yann LeCun, Sumit Chopra, Raia Hadsell, M Ranzato, and Fujie Huang.
\newblock A tutorial on energy-based learning.
\newblock \emph{Predicting Structured Data}, 1\penalty0 (0), 2006.

\bibitem[Lee and Park(2011)]{lee2011fuzzy}
Joonwhoan Lee and EunJong Park.
\newblock Fuzzy similarity-based emotional classification of color images.
\newblock \emph{IEEE Transactions on Multimedia}, 13\penalty0 (5):\penalty0
  1031--1039, 2011.

\bibitem[Liao et~al.(2022)Liao, Hu, Yang, and Rosenhahn]{liao2022text}
Wentong Liao, Kai Hu, Michael~Ying Yang, and Bodo Rosenhahn.
\newblock Text to image generation with semantic-spatial aware gan.
\newblock In \emph{Proceedings of the IEEE/CVF Conference on Computer Vision
  and Pattern Recognition}, pages 18187--18196, 2022.

\bibitem[Liu et~al.(2018)Liu, Jiang, Pei, and Liu]{liu2018emotional}
Da Liu, Yaxi Jiang, Min Pei, and Shiguang Liu.
\newblock Emotional image color transfer via deep learning.
\newblock \emph{Pattern Recognition Letters}, 110:\penalty0 16--22, 2018.

\bibitem[Machajdik and Hanbury(2010)]{machajdik2010affective}
Jana Machajdik and Allan Hanbury.
\newblock Affective image classification using features inspired by psychology
  and art theory.
\newblock In \emph{Proceedings of the 18th ACM International Conference on
  Multimedia}, pages 83--92, 2010.

\bibitem[Mohammad and Kiritchenko(2018)]{mohammad2018wikiart}
Saif Mohammad and Svetlana Kiritchenko.
\newblock Wikiart emotions: An annotated dataset of emotions evoked by art.
\newblock In \emph{Proceedings of the eleventh International Conference on
  Language Resources and Evaluation}, 2018.

\bibitem[Nichol et~al.(2021)Nichol, Dhariwal, Ramesh, Shyam, Mishkin, McGrew,
  Sutskever, and Chen]{nichol2021glide}
Alex Nichol, Prafulla Dhariwal, Aditya Ramesh, Pranav Shyam, Pamela Mishkin,
  Bob McGrew, Ilya Sutskever, and Mark Chen.
\newblock Glide: Towards photorealistic image generation and editing with
  text-guided diffusion models.
\newblock \emph{arXiv preprint arXiv:2112.10741}, 2021.

\bibitem[Oskarsson(2021)]{oskarsson2021robust}
Magnus Oskarsson.
\newblock Robust image-to-image color transfer using optimal inlier
  maximization.
\newblock In \emph{Proceedings of the IEEE/CVF Conference on Computer Vision
  and Pattern Recognition}, pages 786--795, 2021.

\bibitem[Peng et~al.(2015)Peng, Chen, Sadovnik, and Gallagher]{peng2015mixed}
Kuan-Chuan Peng, Tsuhan Chen, Amir Sadovnik, and Andrew~C Gallagher.
\newblock A mixed bag of emotions: Model, predict, and transfer emotion
  distributions.
\newblock In \emph{Proceedings of the IEEE/CVF Conference on Computer Vision
  and Pattern Recognition}, pages 860--868, 2015.

\bibitem[Radford et~al.(2021)Radford, Kim, Hallacy, Ramesh, Goh, Agarwal,
  Sastry, Askell, Mishkin, Clark, et~al.]{radford2021learning}
Alec Radford, Jong~Wook Kim, Chris Hallacy, Aditya Ramesh, Gabriel Goh,
  Sandhini Agarwal, Girish Sastry, Amanda Askell, Pamela Mishkin, Jack Clark,
  et~al.
\newblock Learning transferable visual models from natural language
  supervision.
\newblock In \emph{International Conference on Machine Learning}, pages
  8748--8763, 2021.

\bibitem[Ramesh et~al.(2022)Ramesh, Dhariwal, Nichol, Chu, and
  Chen]{ramesh2022hierarchical}
Aditya Ramesh, Prafulla Dhariwal, Alex Nichol, Casey Chu, and Mark Chen.
\newblock Hierarchical text-conditional image generation with clip latents.
\newblock \emph{arXiv preprint arXiv:2204.06125}, 1\penalty0 (2):\penalty0 3,
  2022.

\bibitem[Rangwani et~al.(2023)Rangwani, Bansal, Sharma, Karmali, Jampani, and
  Babu]{rangwani2023noisytwins}
Harsh Rangwani, Lavish Bansal, Kartik Sharma, Tejan Karmali, Varun Jampani, and
  R~Venkatesh Babu.
\newblock Noisytwins: Class-consistent and diverse image generation through
  stylegans.
\newblock In \emph{Proceedings of the IEEE/CVF Conference on Computer Vision
  and Pattern Recognition}, pages 5987--5996, 2023.

\bibitem[Rao et~al.(2016)Rao, Li, and Xu]{rao2016learning}
Tianrong Rao, Xiaoxu Li, and Min Xu.
\newblock Learning multi-level deep representations for image emotion
  classification.
\newblock \emph{Neural Processing Letters}, pages 1--19, 2016.

\bibitem[Rao et~al.(2020)Rao, Li, and Xu]{rao2020learning}
Tianrong Rao, Xiaoxu Li, and Min Xu.
\newblock Learning multi-level deep representations for image emotion
  classification.
\newblock \emph{Neural Processing Letters}, 51:\penalty0 2043--2061, 2020.

\bibitem[Reinhard et~al.(2001)Reinhard, Adhikhmin, Gooch, and
  Shirley]{reinhard2001color}
Erik Reinhard, Michael Adhikhmin, Bruce Gooch, and Peter Shirley.
\newblock Color transfer between images.
\newblock \emph{IEEE Computer Graphics and Applications}, 21\penalty0
  (5):\penalty0 34--41, 2001.

\bibitem[Rezende and Mohamed(2015)]{rezende2015variational}
Danilo Rezende and Shakir Mohamed.
\newblock Variational inference with normalizing flows.
\newblock In \emph{International Conference on Machine Learning}, pages
  1530--1538, 2015.

\bibitem[Rombach et~al.(2022)Rombach, Blattmann, Lorenz, Esser, and
  Ommer]{rombach2022high}
Robin Rombach, Andreas Blattmann, Dominik Lorenz, Patrick Esser, and Bj{\"o}rn
  Ommer.
\newblock High-resolution image synthesis with latent diffusion models.
\newblock In \emph{Proceedings of the IEEE/CVF Conference on Computer Vision
  and Pattern Recognition}, pages 10684--10695, 2022.

\bibitem[Ruiz et~al.(2023)Ruiz, Li, Jampani, Pritch, Rubinstein, and
  Aberman]{ruiz2023dreambooth}
Nataniel Ruiz, Yuanzhen Li, Varun Jampani, Yael Pritch, Michael Rubinstein, and
  Kfir Aberman.
\newblock Dreambooth: Fine tuning text-to-image diffusion models for
  subject-driven generation.
\newblock In \emph{Proceedings of the IEEE/CVF Conference on Computer Vision
  and Pattern Recognition}, pages 22500--22510, 2023.

\bibitem[Saharia et~al.(2022)Saharia, Chan, Saxena, Li, Whang, Denton,
  Ghasemipour, Gontijo~Lopes, Karagol~Ayan, Salimans,
  et~al.]{saharia2022photorealistic}
Chitwan Saharia, William Chan, Saurabh Saxena, Lala Li, Jay Whang, Emily~L
  Denton, Kamyar Ghasemipour, Raphael Gontijo~Lopes, Burcu Karagol~Ayan, Tim
  Salimans, et~al.
\newblock Photorealistic text-to-image diffusion models with deep language
  understanding.
\newblock \emph{Advances in Neural Information Processing Systems},
  35:\penalty0 36479--36494, 2022.

\bibitem[Sun et~al.(2023)Sun, Jia, Wu, Ye, and Xing]{sun2023msnet}
Shikun Sun, Jia Jia, Haozhe Wu, Zijie Ye, and Junliang Xing.
\newblock Msnet: A deep architecture using multi-sentiment semantics for
  sentiment-aware image style transfer.
\newblock In \emph{ICASSP 2023-2023 IEEE International Conference on Acoustics,
  Speech and Signal Processing}, pages 1--5. IEEE, 2023.

\bibitem[Tan et~al.(2021)Tan, Chai, Chen, Liao, Chu, Liu, Hua, and
  Yu]{tan2021diverse}
Zhentao Tan, Menglei Chai, Dongdong Chen, Jing Liao, Qi Chu, Bin Liu, Gang Hua,
  and Nenghai Yu.
\newblock Diverse semantic image synthesis via probability distribution
  modeling.
\newblock In \emph{Proceedings of the IEEE/CVF Conference on Computer Vision
  and Pattern Recognition}, pages 7962--7971, 2021.

\bibitem[Tevet et~al.(2022)Tevet, Gordon, Hertz, Bermano, and
  Cohen-Or]{tevet2022motionclip}
Guy Tevet, Brian Gordon, Amir Hertz, Amit~H Bermano, and Daniel Cohen-Or.
\newblock Motionclip: Exposing human motion generation to clip space.
\newblock In \emph{European Conference on Computer Vision}, pages 358--374.
  Springer, 2022.

\bibitem[Wang et~al.(2013)Wang, Jia, and Cai]{wang2013affective}
Xiaohui Wang, Jia Jia, and Lianhong Cai.
\newblock Affective image adjustment with a single word.
\newblock \emph{The Visual Computer}, 29:\penalty0 1121--1133, 2013.

\bibitem[Wang et~al.(2023)Wang, Zhao, and Xing]{wang2023stylediffusion}
Zhizhong Wang, Lei Zhao, and Wei Xing.
\newblock Stylediffusion: Controllable disentangled style transfer via
  diffusion models.
\newblock In \emph{Proceedings of the IEEE/CVF International Conference on
  Computer Vision}, pages 7677--7689, 2023.

\bibitem[Wei et~al.(2023)Wei, Zhang, Ji, Bai, Zhang, and Zuo]{wei2023elite}
Yuxiang Wei, Yabo Zhang, Zhilong Ji, Jinfeng Bai, Lei Zhang, and Wangmeng Zuo.
\newblock Elite: Encoding visual concepts into textual embeddings for
  customized text-to-image generation.
\newblock \emph{arXiv preprint arXiv:2302.13848}, 2023.

\bibitem[Weng et~al.(2023)Weng, Zhang, Chang, Wang, Li, and
  Shi]{weng2023affective}
Shuchen Weng, Peixuan Zhang, Zheng Chang, Xinlong Wang, Si Li, and Boxin Shi.
\newblock Affective image filter: Reflecting emotions from text to images.
\newblock In \emph{Proceedings of the IEEE/CVF Conference on Computer Vision
  and Pattern Recognition}, pages 10810--10819, 2023.

\bibitem[Yadollahi et~al.(2017)Yadollahi, Shahraki, and
  Zaiane]{yadollahi2017current}
Ali Yadollahi, Ameneh~Gholipour Shahraki, and Osmar~R Zaiane.
\newblock Current state of text sentiment analysis from opinion to emotion
  mining.
\newblock \emph{ACM Computing Surveys (CSUR)}, 50\penalty0 (2):\penalty0 1--33,
  2017.

\bibitem[Yang and Peng(2008)]{yang2008automatic}
Chuan-Kai Yang and Li-Kai Peng.
\newblock Automatic mood-transferring between color images.
\newblock \emph{IEEE Computer Graphics and Applications}, 28\penalty0
  (2):\penalty0 52--61, 2008.

\bibitem[Yang et~al.(2018)Yang, She, Lai, Rosin, and Yang]{yang2018weakly}
Jufeng Yang, Dongyu She, Yu-Kun Lai, Paul~L Rosin, and Ming-Hsuan Yang.
\newblock Weakly supervised coupled networks for visual sentiment analysis.
\newblock In \emph{Proceedings of the IEEE/CVF Conference on Computer Vision
  and Pattern Recognition}, pages 7584--7592, 2018.

\bibitem[Yang et~al.(2021{\natexlab{a}})Yang, Gao, Li, Wang, and
  Ding]{yang2021solver}
Jingyuan Yang, Xinbo Gao, Leida Li, Xiumei Wang, and Jinshan Ding.
\newblock Solver: Scene-object interrelated visual emotion reasoning network.
\newblock \emph{IEEE Transactions on Image Processing}, 30:\penalty0
  8686--8701, 2021{\natexlab{a}}.

\bibitem[Yang et~al.(2021{\natexlab{b}})Yang, Li, Wang, Ding, and
  Gao]{yang2021stimuli}
Jingyuan Yang, Jie Li, Xiumei Wang, Yuxuan Ding, and Xinbo Gao.
\newblock Stimuli-aware visual emotion analysis.
\newblock \emph{IEEE Transactions on Image Processing}, 30:\penalty0
  7432--7445, 2021{\natexlab{b}}.

\bibitem[Yang et~al.(2023)Yang, Huang, Ding, Lischinski, Cohen-Or, and
  Huang]{yang2023emoset}
Jingyuan Yang, Qirui Huang, Tingting Ding, Dani Lischinski, Danny Cohen-Or, and
  Hui Huang.
\newblock Emoset: A large-scale visual emotion dataset with rich attributes.
\newblock In \emph{Proceedings of the IEEE/CVF International Conference on
  Computer Vision}, pages 20383--20394, 2023.

\bibitem[Yang et~al.(2022)Yang, Jiang, Liu, and Loy]{yang2022pastiche}
Shuai Yang, Liming Jiang, Ziwei Liu, and Chen~Change Loy.
\newblock Pastiche master: Exemplar-based high-resolution portrait style
  transfer.
\newblock In \emph{Proceedings of the IEEE/CVF Conference on Computer Vision
  and Pattern Recognition}, pages 7693--7702, 2022.

\bibitem[Zhang and Peng(2018)]{zhang2018stacking}
Chenrui Zhang and Yuxin Peng.
\newblock Stacking vae and gan for context-aware text-to-image generation.
\newblock In \emph{2018 IEEE Fourth International Conference on Multimedia Big
  Data (BigMM)}, pages 1--5. IEEE, 2018.

\bibitem[Zhang et~al.(2023)Zhang, Rao, and Agrawala]{zhang2023adding}
Lvmin Zhang, Anyi Rao, and Maneesh Agrawala.
\newblock Adding conditional control to text-to-image diffusion models.
\newblock In \emph{Proceedings of the IEEE/CVF International Conference on
  Computer Vision}, pages 3836--3847, 2023.

\bibitem[Zhang et~al.(2018)Zhang, Isola, Efros, Shechtman, and
  Wang]{zhang2018unreasonable}
Richard Zhang, Phillip Isola, Alexei~A Efros, Eli Shechtman, and Oliver Wang.
\newblock The unreasonable effectiveness of deep features as a perceptual
  metric.
\newblock In \emph{Proceedings of the IEEE/CVF Conference on Computer Vision
  and Pattern Recognition}, pages 586--595, 2018.

\bibitem[Zhang et~al.(2019)Zhang, He, and Lu]{zhang2019exploring}
Wei Zhang, Xuanyu He, and Weizhi Lu.
\newblock Exploring discriminative representations for image emotion
  recognition with cnns.
\newblock \emph{IEEE Transactions on Multimedia}, 22\penalty0 (2):\penalty0
  515--523, 2019.

\bibitem[Zhu et~al.(2019)Zhu, Pan, Chen, and Yang]{zhu2019dm}
Minfeng Zhu, Pingbo Pan, Wei Chen, and Yi Yang.
\newblock Dm-gan: Dynamic memory generative adversarial networks for
  text-to-image synthesis.
\newblock In \emph{Proceedings of the IEEE/CVF Conference on Computer Vision
  and Pattern Recognition}, pages 5802--5810, 2019.

\bibitem[Zhu et~al.(2023)Zhu, Qing, Chen, and Xu]{zhu2023emotional}
Siqi Zhu, Chunmei Qing, Canqiang Chen, and Xiangmin Xu.
\newblock Emotional generative adversarial network for image emotion transfer.
\newblock \emph{Expert Systems with Applications}, 216:\penalty0 119485, 2023.

\end{thebibliography}
}


\end{document}